\documentclass[letterpaper, 10 pt, conference]{ieeeconf}  

\usepackage{amsmath,amssymb,amsfonts}
\usepackage{balance}
\usepackage{bm}
\usepackage{hyperref}
\usepackage{color}
\usepackage{cite}
\usepackage[font=footnotesize]{caption}
\usepackage[table,xcdraw]{xcolor}
\usepackage{tabularx}
\usepackage{booktabs}
\usepackage{graphicx}
\usepackage{multirow}
\usepackage{tikz}
\usepackage{tikz-3dplot}
\usepackage{pgfplots}
\usepackage{units}
\usepackage{siunitx}
\usepackage{pifont,hologo}
\usetikzlibrary{bending, external, patterns,shapes.arrows, matrix}
\usepgfplotslibrary{groupplots,dateplot}

\IEEEoverridecommandlockouts                             
\DeclareUnicodeCharacter{2212}{−}

\pgfplotsset{compat=newest}

\NewDocumentCommand{\evalat}{sO{\big}mm}{%
  \IfBooleanTF{#1}
   {\mleft. #3 \mright|_{#4}}
   {#3#2|_{#4}}%
}

\newcommand\mypara[1]{\vspace{3pt}\noindent\textbf{#1.}}

\newcommand{\wfr}[0]{\ensuremath{\mathcal{W}}} %
\newcommand{\bfr}[0]{\ensuremath{\mathcal{B}}} %

\newcommand{\rotateRPY}[3]%
{   \pgfmathsetmacro{\rollangle}{#1}
    \pgfmathsetmacro{\pitchangle}{#2}
    \pgfmathsetmacro{\yawangle}{#3}

    \pgfmathsetmacro{\newxx}{cos(\yawangle)*cos(\pitchangle)}
    \pgfmathsetmacro{\newxy}{sin(\yawangle)*cos(\pitchangle)}
    \pgfmathsetmacro{\newxz}{-sin(\pitchangle)}
    \path (\newxx,\newxy,\newxz);
    \pgfgetlastxy{\nxx}{\nxy};

    \pgfmathsetmacro{\newyx}{cos(\yawangle)*sin(\pitchangle)*sin(\rollangle)-sin(\yawangle)*cos(\rollangle)}
    \pgfmathsetmacro{\newyy}{sin(\yawangle)*sin(\pitchangle)*sin(\rollangle)+ cos(\yawangle)*cos(\rollangle)}
    \pgfmathsetmacro{\newyz}{cos(\pitchangle)*sin(\rollangle)}
    \path (\newyx,\newyy,\newyz);
    \pgfgetlastxy{\nyx}{\nyy};

    \pgfmathsetmacro{\newzx}{cos(\yawangle)*sin(\pitchangle)*cos(\rollangle)+ sin(\yawangle)*sin(\rollangle)}
    \pgfmathsetmacro{\newzy}{sin(\yawangle)*sin(\pitchangle)*cos(\rollangle)-cos(\yawangle)*sin(\rollangle)}
    \pgfmathsetmacro{\newzz}{cos(\pitchangle)*cos(\rollangle)}
    \path (\newzx,\newzy,\newzz);
    \pgfgetlastxy{\nzx}{\nzy};
}

\newcolumntype{C}{>{\centering\arraybackslash}X}

\newcommand\clearrow{\global\let\rowmac\relax}
\clearrow

\newcommand{\mat}[1]{\begin{bmatrix} #1 \end{bmatrix}}

\newcommand{\rom}[1]{\expandafter{\romannumeral #1\relax}}

\newcommand{\xmark}{\ding{55}}%

\newcommand{\hide}[1]{}

\newcommand{\rev}[1]{{\color{black}#1}}
\newcommand\icra[1]{\textcolor{black}{#1}}

\definecolor{c1}{HTML}{D9B08C}
\definecolor{c2}{HTML}{FFCB9A}
\definecolor{c3}{HTML}{D1E8FF}
\definecolor{c4}{HTML}{DEF2F1}

\definecolor{somegray}{rgb}{0.5, 0.5, 0.5}
\newcommand{\darkgrayed}[1]{\textcolor{somegray}{#1}}
\makeatletter
\newcommand*\titleheader[1]{\gdef\@titleheader{#1}}
\AtBeginDocument{%
  \let\st@red@title\@title
  \def\@title{%
    \vskip-3em
    \bgroup\normalfont\large\centering\@titleheader\par\egroup
    \vskip1.1em\st@red@title}
}
\makeatother

\titleheader{\darkgrayed{This paper has been accepted for publication at the \\ IEEE International Conference on Robotics and Automation (ICRA), Philadelphia, 2022. \copyright IEEE}}

\title{\bf\LARGE A Benchmark Comparison of Learned Control Policies\\for Agile Quadrotor Flight}

\begin{document}

\definecolor{color1}{rgb}{0, 0.4470, 0.7410}
\definecolor{color2}{rgb}{0.8500, 0.3250, 0.0980}
\definecolor{color3}{rgb}{0.9290, 0.6940, 0.1250}
\definecolor{color4}{rgb}{0.4660, 0.6740, 0.1880}

\makeatletter
\g@addto@macro\@maketitle{
  \captionsetup{type=figure}\setcounter{figure}{0}
  \def\mycolspace{1.2mm}
    \centering
    \begin{tikzpicture}[>=stealth]
\tikzstyle{every node}=[font=\scriptsize]

\tikzset{
  box/.style={
    draw,
	inner sep=3pt,
	outer sep=0pt,
	align=center,
	minimum height = 0.4cm,
	minimum width = 1cm,
  }  
}

\tikzset{
  label/.style={
	above,
	align=center
  }  
}

\tikzset{
  labelend/.style={
	above,
	near end,
	anchor=south
  }  
}

\tikzset{
  labelstart/.style={
	above,
	pos=0,
	anchor=south west,
  }  
}

\def\dx{1.5cm}
\def\dy{1.5cm}
\def\halfWidth{2.5cm}

\begin{scope}[xshift=12cm]
\coordinate (srt_left) at (-\halfWidth,0);
\coordinate (srt_right) at (1cm,0);
\path (srt_left)  -- node [above=5mm, midway] {\bf SRT} (srt_right);
\node[box, fill=lightgray] at (-\dx,0) (srt_pol) {$\pi_\text{SRT}$};
\node[box] at(0,0) (srt_delay) {Delay};
\node[box] at(-\dx, -\dy) (srt_system) {$\bf{\dot{x}} = f(\bf x)$};
\draw [->] (srt_pol) -- node [label] {$\tau_i$} (srt_delay);
\draw [->] (srt_delay) -- (srt_delay -| srt_right) |-  (srt_system);
\draw [->] (srt_system) -| (srt_pol -| srt_left) |-  node [labelend] {$\mathbf o_t$} (srt_pol);
\node [anchor=south west, inner sep=0mm, outer sep=0mm] (srt_text) at ([xshift=3mm, yshift=5mm] srt_system.east) {$\Delta t=\unit[1]{ms}$};
\draw [dashed] ([xshift=-2mm, yshift=-4mm]srt_system.south west) rectangle ([xshift=1mm, yshift=1mm]srt_text.north east);
\end{scope}

\begin{scope}[xshift=0cm]
\coordinate (lv_left) at (-\halfWidth,0);
\coordinate (lv_right) at (\halfWidth,0);
\path (lv_left)  -- node [above=5mm, midway] {\bf LV} (lv_right);
\node[box,fill=lightgray] at (-\dx,0) (lv_pol) {$\pi_\text{LV}$};
\node[box] at(0,0) (lv_delay) {Delay};
\node[box] at([xshift=-1mm]\dx,-\dy) (lv_ctrl) {Ctrl};
\node[box] at (-\dx,-\dy) (lv_system) {$\bf{\dot{x}} = f(\bf x)$};
\node[box] at (0,-\dy) (lv_pd) {PD};
\draw [->] (lv_pol) -- node [midway, align=center, text width = 0.5cm, inner sep = 0]   {$\mathbf v$ \\[2pt] $\omega_z$}  (lv_delay);
\draw [->] (lv_delay) -- (lv_right) |- (lv_ctrl);
\draw [->] (lv_ctrl) -- (lv_pd);
\draw [->] (lv_pd) -- node [label] {$\tau_i$} (lv_system);
\draw [->] (lv_system) -| (lv_pol -| lv_left) |-  node [labelend] {$\mathbf{o}_t$} (lv_pol);
\draw [->] (lv_system) -- ([yshift=1.5mm]lv_system.north) -| node [above, very near start] {$\mathbf x$} (lv_ctrl);
\node [anchor=south west, inner sep=0mm, outer sep=0mm] (lv_text) at ([xshift=3mm, yshift=5mm] lv_pd.east) {$\Delta t=\unit[1]{ms}$};
\draw [dashed] ([xshift=-2mm, yshift=-4mm]lv_system.south west) rectangle ([xshift=1mm, yshift=1mm]lv_text.north east);
\draw [->] (lv_system) |- node [near end, above=-.7mm] {IMU} ([yshift=-3mm]lv_pd.south) -- (lv_pd);
\end{scope}

\begin{scope}[xshift=6.75cm]
\coordinate (ctrb_left) at (-\halfWidth,0);
\coordinate (ctrb_right) at (1cm,0);
\path (ctrb_left)  -- node [above=5mm, midway] {\bf CTBR} (ctrb_right);
\node[box,fill=lightgray] at (-\dx,0) (ctrb_pol) {$\pi_\text{CTBR}$};
\node[box] at(0,0) (ctrb_delay) {Delay};
\node[box] at (-\dx,-\dy) (ctrb_system) {$\bf{\dot{x}} = f(\bf x)$};
\node[box] at (0,-\dy) (ctrb_pd) {PD};
\draw [->] (ctrb_pol) -- node [midway, align=center, text width = 0.5cm, inner sep = 0] {$c$ \\[2pt] $\boldsymbol\omega$}  (ctrb_delay);
\draw [->] (ctrb_delay) -- (ctrb_delay -| ctrb_right) |- (ctrb_pd);
\draw [->] (ctrb_pd) -- node [label] {$\tau_i$} (ctrb_system);
\draw [->] (ctrb_system) -| (ctrb_pol -| ctrb_left) |-  node [labelend] {$\mathbf o_t$} (ctrb_pol);
\node [anchor=south east, inner sep=0mm, outer sep=0mm] (ctrb_text) at ([xshift=2mm, yshift=5mm] ctrb_pd.east) {$\Delta t=\unit[1]{ms}$};
\draw [->] (ctrb_system) |- node [near end, above=-0.7mm] {IMU} ([yshift=-3mm]ctrb_pd.south) -- (ctrb_pd);
\draw [dashed] ([xshift=-2mm, yshift=-4mm]ctrb_system.south west) rectangle ([xshift=1mm, yshift=1mm]ctrb_text.north east) ;
\end{scope}

\end{tikzpicture}
    
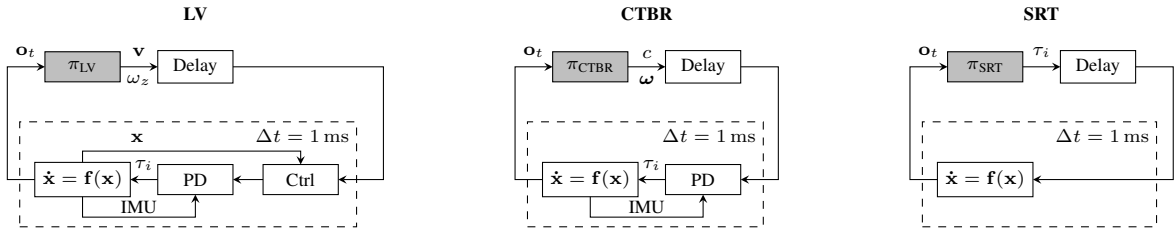
\captionof{figure}{In this paper, we compare three different classes of control policies for the task of agile quadrotor flight. From left to right: policies \rev{specifying} desired linear velocities (LV) (they rely on a control stack that maps the output velocities to individual rotor thrusts), policies \rev{commanding} collective thrust and bodyrates (CTBR) (they rely on a low-level controller that maps the output bodyrates to individual rotor thrusts), policies directly \rev{outputting} single-rotor thrust (SRT).
	\label{fig:overview}
	\vspace{-9pt}
}
}

\author{Elia Kaufmann,
        Leonard Bauersfeld,
        Davide Scaramuzza
         \thanks{The authors are with the Robotics and Perception Group, Dep. of Informatics, University of Zurich, and Dep. of Neuroinformatics, University of Zurich and ETH Zurich, Switzerland (\protect\url{https://rpg.ifi.uzh.ch}).
        This work was supported by the National Centre of Competence in Research (NCCR) Robotics through the Swiss National Science Foundation (SNSF) and the European Union’s Horizon 2020 Research and Innovation Programme under grant agreement No.~871479 (AERIAL-CORE) and the European Research Council (ERC) under grant agreement No.~864042 (AGILEFLIGHT).
        }
}

\maketitle

\begin{abstract}
Quadrotors are highly nonlinear \rev{dynamical} systems that require carefully tuned controllers to be pushed to their physical limits. 
Recently, learning-based control policies have been proposed for quadrotors, 
as they would potentially allow learning direct mappings from high-dimensional raw sensory observations to actions. 
Due to sample inefficiency, training such learned controllers on the real platform is impractical or even impossible. 
Training in simulation is attractive but requires to transfer policies between domains, which demands trained policies to be robust to such domain gap. 
In this work, we make two contributions: 
(i) we perform the first benchmark comparison of existing learned control policies for agile quadrotor flight and show that training a control policy that \rev{commands} body-rates and thrust results in more robust sim-to-real transfer compared to a policy that directly \rev{specifies} individual rotor thrusts,
(ii)~we demonstrate for the first time that such a control policy trained via deep reinforcement learning can control a quadrotor in real-world experiments at speeds over \unit[45]{km/h}.

\end{abstract}

\section*{Supplementary Material}
A narrated video illustrating our findings is available at
\url{https://youtu.be/zqdfVq2uWUA}

\section{Introduction}\label{sec: introduction}

Agile quadrotor flight is a challenging problem that requires fast and accurate control strategies.
In recent years, numerous learning-based controllers have been proposed for quadrotors. 
In contrast to their traditional counterparts, learned control policies have the potential to directly map sensory information to actions, alleviating the need for explicit state estimation~\cite{zhang2016learning, kaufmann2020deep, levine2016end, Loquercio2021Science}.

Prior work has proposed learned control policies that make use of various control input modalities to the underlying platform: while some directly \rev{specify} motor commands~\cite{zhang2016learning, hwangbo2017control, molchanov2019sim}, others, instead, \rev{output} desired collective thrust and bodyrates~\cite{kaufmann2020deep, muller2018teaching} (that are then executed by a low-level controller), or velocity commands~\cite{giusti2016machine, loquercio2018dronet} (that are then executed by a control stack), or even a sequence of future waypoints~\cite{Loquercio2021Science}.
Most published approaches do not justify their choice of control input. 
This renders performance comparisons among them and, thus, scientific progress difficult.

Due to the high sample complexity of learning-based policies, they are often trained in simulation, which then requires transferring the policy from simulation to the real world. 
This transfer between domains is known to be hard and is typically approached by increasing the simulation fidelity~\cite{tan2018sim, bauersfeld2021neurobem}, by randomization of dynamics~\cite{molchanov2019sim, andrychowicz2020learning} or rendering properties~\cite{sadeghi2017cadrl, tobin2017domain} at training time, or by abstraction of the policy inputs~\cite{kaufmann2020deep, Loquercio2021Science}.
Apart from simulation enhancements and input abstractions, also 
\rev{the choice of action space} 
of the learned policy itself can facilitate transfer.
Policies that \rev{generate} high-level commands, such as desired linear velocity or future waypoints~\cite{Loquercio2021Science}, have a reduced simulation to reality gap, as they abstract the task of flying by relying on an existing underlying control stack. 
However, while facilitating transfer, such abstractions also constrain the maneuverability of the platform. 
Approaches that do not rely on such abstractions (like those \rev{specifying} collective thrust and body rates or even single-rotor-thrust commands) can potentially execute much more agile maneuvers, but have so far only been shown for near-hover trajectories~\cite{molchanov2019sim} or require a dedicated policy for each maneuver~\cite{kaufmann2020deep}.

In this paper, we compare and benchmark learned control policies \rev{with respect to their choice of action space.} %
Specifically, we compare them in terms of peak performance in case of perfect model identification, as well as in terms of their transferability to a new platform with possibly different dynamics properties. 
We compare the learned policies with respect to their flight performance, which we characterize by the \icra{average} tracking error on a set of predefined trajectories.

Our experiments, performed both in simulation and on a real quadrotor platform, show that control policies that \rev{command} collective thrust and bodyrates are more robust to changes in the dynamics of the platform without compromising agility.
Additionally, compared to high-level action parameterizations, specifying collective thrust and bodyrates allows performing significantly more agile maneuvers. 

Finally, we demonstrate the first learning-based controller, trained via deep reinforcement learning, that is able to perform previously unseen agile maneuvers on a real quadrotor flying at speeds over \SI{45}{\kilo\meter\per\hour}.
The policy is trained purely in simulation and transferred to the real platform without any fine-tuning.

\section{Related Work}
\label{sec: related work}
In this section, we give an overview of the related work for learning-based quadrotor control while focusing on the choice of action space.
\rev{While there exists a comparison of action spaces of learned policies for 2D locomotion~\cite{peng2017learning}, such analysis is still lacking in the aerial robotics community.}
In the following,
we group learned control strategies according to \rev{their action space} into a)~Linear Velocity Commands~(LV), b)~Collective Thrust and Bodyrates~(CTBR), and c)~Single Rotor Thrusts~(SRT).

\mypara{Linear Velocity}
Control policies \rev{specifying} high-level commands, often in the form of receding-horizon waypoints or velocity commands, have been proposed for a variety of tasks, such as forest trail navigation~\cite{giusti2016machine}, navigation in city streets~\cite{loquercio2018dronet} and indoor environments~\cite{sadeghi2017cadrl}, or even drone racing~\cite{kaufmann2018deep}. 
Recently, \cite{belkhale2021model} have used model-based meta reinforcement learning to \rev{generate} velocity commands that adapt to unknown payloads.
While these approaches have been successfully deployed in the real world, only~\cite{kaufmann2018deep} achieved flight speeds beyond \SI{3}{\meter\per\second}, while the other policies result in near-hover flight.
As the control policy does not take into account the dynamic constraints of the platform, it can be easily transferred, but does not exploit the platform's full dynamic capabilities. 
Furthermore, such approaches rely on an existing underlying control stack, which itself is dependent on high-quality state estimation. 

\mypara{Collective Thrust and Bodyrates} Compared to \rev{specifying} linear velocity commands, controlling collective thrust and bodyrates has been shown to allow performing significantly more aggressive maneuvers. 
In~\cite{muller2018teaching}, a racing policy directly maps image observations to collective thrust and bodyrate commands. Although the policy successfully races through challenging race tracks in simulation, it is not deployed on a real platform.
\rev{In~\cite{shi2019neural}, the authors propose combining a classical controller with a learned residual command to correct for aerodynamic disturbances such as ground effect during near-hover flight.}
In~\cite{kaufmann2020deep}, the authors use privileged learning to imitate a model predictive controller~(MPC) to perform acrobatic maneuvers. 
While this approach successfully showed acrobatic flight on a real platform, it was constrained to a single maneuver and required a separate policy for each trajectory.
In contrast to \rev{generating} high-level commands, \rev{specifying} collective thrust and bodyrates does not necessitate estimation of the full state of the platform, but only requires inertial measurements to perform feedback control on the bodyrates. 
This information is readily available at high frequency in today's flight controllers, rendering collective thrust and bodyrates the preferred control input modality for professional human pilots. 

\mypara{Single-Rotor Thrusts} There are several works that propose to directly learn to control individual rotor thrusts~\cite{hwangbo2017control, molchanov2019sim, song2021autonomous, zhang2016learning, lambert2019low, pi2020low, pi2021robust}. 
As this control input does not require any additional control loop, it provides direct access to the actuators and as a result correctly represents the true actuation limits of the platform. 
It constitutes the most versatile control input investigated in this work.
In~\cite{hwangbo2017control, molchanov2019sim}, the authors train a policy to map state observations directly to desired individual rotor thrusts. 
While~\cite{hwangbo2017control} required a PID controller at data collection time to facilitate training, \cite{molchanov2019sim}~demonstrated training of a stabilizing quadrotor control policy from scratch in simulation and deployment on multiple real platforms. 
\cite{song2021autonomous}~trains a policy for autonomous drone racing. Their approach demonstrates competitive racing performance in simulation, but is not deployed on a real quadrotor.
In~\cite{zhang2016learning}, the authors train a policy to perform obstacle avoidance using guided policy search by imitating an MPC controller that has access to privileged information about the environment. 
One of the few works that does not rely on simulated data for training is presented in~\cite{lambert2019low}, where the authors propose an approach based on deep model-based reinforcement learning to train a hovering policy for the \textit{Crazyflie} quadrotor. The trained policy managed to control the real platform in hover for \SI{6}{\second} before crashing. 
A position controller is trained via reinforcement learning in~\cite{pi2020low} and extended in~\cite{pi2021robust} to be robust against external disturbances such as wind. 
In~\cite{koch2019reinforcement}, the authors train an attitude controller via deep reinforcement learning. They argue that their approach provides a better flight performance compared to a PID controller, while still being computationally lightweight.
\rev{Although this method outputs} individual rotor thrusts, it is still dependent on a higher-level controller that produces attitude setpoints to achieve stable flight.

While some of these works show successful deployment of their policies in the real world, none achieved agile flight, only reaching maximum speeds below \SI{4}{\meter\per\second}.

\section{Quadrotor Dynamics}
\label{sec:preliminaries}

To train a control policy for agile flight, we implement the quadrotor dynamics as an environment in TensorFlow~Agents\footnote{\url{https://github.com/tensorflow/agents}}. 
The following section gives a brief overview of the dynamics implemented in the simulator. 

\subsection{Notation}
Scalars are denoted in non-bold~$[s, S]$, vectors in lowercase bold~$\bm{v}$, and matrices in uppercase bold~$\bm{M}$.
World $\wfr$ and Body $\bfr$ frames are defined with orthonormal basis i.e. $\{\bm{x}_\wfr, \bm{y}_\wfr, \bm{z}_\wfr\}$.
The frame $\bfr$ is located at the center of mass of the quadrotor.
A vector from coordinate $\bm{p}_1$ to $\bm{p}_2$ expressed in the $\wfr$ frame is written as: ${}_\wfr\bm{v}_{12}$.
If the vector's origin coincides with the frame it is described in, the frame index is dropped, e.g. the quadrotor position is denoted as $\bm{p}_{\wfr\bfr}$.
Unit quaternions $\bm{q} = (q_w, q_x, q_y, q_z)$ with $\|\bm{q}\| = 1$ are used to represent orientations, such as the attitude state of the quadrotor body $\bm{q}_{\wfr\bfr}$.
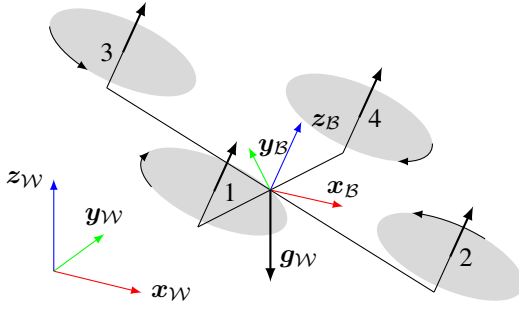
\begin{figure}
\centering
\tdplotsetmaincoords{65}{30}
\begin{tikzpicture}[xscale=0.9,yscale=0.9,tdplot_main_coords, >=latex]
\tikzset{RPY/.style={x={(\nxx,\nxy)},y={(\nyx,\nyy)},z={(\nzx,\nzy)}}}
\def\propz{0.7}
\def\propx{3.3}
\def\propy{2.7}
\def\r{1.3}
\def\l{1.5}
\def\f{1}

\rotateRPY{-10}{20}{0}
\begin{scope}[RPY]

\draw[] (-\propx,0,0) -- (\propx,0,0);
\draw[] (0,-\propy,0) -- (0,\propy,0);

\draw[draw=none, fill=black, opacity=0.15] (0,\propy,\propz) circle (\r);
\draw[draw=none, fill=black, opacity=0.15] (\propx,0,\propz) circle (\r) ;
\draw[draw=none, fill=black, opacity=0.15] (0,-\propy,\propz) circle (\r) ;
\draw[draw=none, fill=black, opacity=0.15] (-\propx,0,\propz) circle (\r) ;
\draw (0,\propy,0) --++ (0,0,\propz) node[right] {4};
\draw (\propx,0,0) --++ (0,0,\propz) node[right] {2};
\draw (0,-\propy,0) --++ (0,0,\propz) node[right] {1};
\draw (-\propx,0,0) --++ (0,0,\propz) node[left] {3};

\draw[thick,->] (0,\propy,\propz) -- ++(0,0,\f);
\draw[thick,->] (\propx,0,\propz) -- ++(0,0,\f);
\draw[thick,->] (0,-\propy,\propz) -- ++(0,0,\f);
\draw[thick,->] (-\propx,0,\propz) -- ++(0,0,\f);	

\def\psi{60}
\def\xa{30}
\def\xb{90}
\def\xc{-120}
\def\xd{210}
\path (\r,\propy,\propz) arc (0:\xa:\r) coordinate (a);
\draw [->] (a) arc (\xa:\xa-\psi:\r) [draw=black] ;

\path (\propx,\r,\propz) arc (90:\xb:\r) coordinate (b);
\draw [->] (b) arc (\xb:\xb+\psi:\r) [draw=black] ;

\path (\r,-\propy,\propz) arc (0:\xc:\r) coordinate (c);
\draw [->] (c) arc (\xc:\xc-\psi:\r) [draw=black] ;

\path (-\propx,\r,\propz)  arc (90:\xd:\r) coordinate (d);
\draw [->] (d) arc (\xd:\xd+\psi:\r) [draw=black] ;

\rotateRPY{0}{0}{45}
\begin{scope}[RPY]
\draw[->,color=red,text=black] (0,0,0) -- ++ (\l,0,0) node[above] {$\bm x_\bfr$};
\draw[->,color=green,text=black] (0,0,0) -- ++ (0,\l,0) node[right] {$\bm y_\bfr$};	
\draw[->,color=blue,text=black] (0,0,0) -- ++ (0,0,\l) node[right] {$\bm z_\bfr$};		
\end{scope}
\end{scope}

\begin{scope}[xshift=-3.2cm, yshift=-1.2cm]
\draw[->,color=red,text=black] (0,0,0) -- ++ (\l,0,0) node[right] {$\bm x_\wfr$};
\draw[->,color=green,text=black] (0,0,0) -- ++ (0,\l,0) node[above] {$\bm y_\wfr$};	
\draw[->,color=blue,text=black] (0,0,0) -- ++ (0,0,\l) node[left] {$\bm z_\wfr$};		
\end{scope}

\draw[thick,->,color=black,text=black] (0,0,0) -- node[right, near end] {$\bm g_\wfr$} ++ (0,0,-1.5*\f);

\end{tikzpicture}
\caption{Diagram of the quadrotor depicting the world and body frames and propeller numbering.}
\label{fig:quad_schematic}
\vspace*{-12pt}
\end{figure}

Finally, full SE3 transformations, such as changing the frame of reference from body to world for a point $\bm{p}_{B1}$, can be described by $_\wfr\bm{p}_{B1} = {}_\wfr\bm{t}_{\wfr\bfr} + \bm{q}_{\wfr\bfr} \odot \bm{p}_{B1}$.
Note the quaternion-vector product is denoted by $\odot$ representing a rotation of the vector by the quaternion as in $\bm{q} \odot \bm{v} = \bm{q} \bm{v} \bar{\bm{q}}$, where $\bar{\bm{q}}$ is the quaternion's conjugate.

\subsection{Quadrotor Dynamics}
The quadrotor is assumed to be a 6 degree-of-freedom rigid body of mass $m$ and diagonal moment of inertia matrix $\bm{J}=\mathrm{diag}(J_x, J_y, J_z)$. Furthermore, the rotational speeds of the four propellers $\Omega_i$ are modeled as first-order system with time constant $k_\text{mot}$ where the commanded motor speeds $\boldsymbol\Omega_\text{cmd}$ are the input.

The state space is thus 17-dimensional and its dynamics can be written as:
\begin{align}
\small
\label{eq:3d_quad_dynamics}
\dot{\bm{x}} =
\begin{bmatrix}
\dot{\bm{p}}_{\wfr\bfr} \\[3pt]
\dot{\bm{q}}_{\wfr\bfr} \\[3pt]
\dot{\bm{v}}_{\wfr\bfr} \\[3pt]
\dot{\boldsymbol\omega}_\bfr \\[3pt]
\dot{\boldsymbol\Omega}
\end{bmatrix} = 
\begin{bmatrix}
\bm{v}_\wfr \\[3pt]
\bm{q}_{\wfr\bfr} \cdot \mat{0 \\ \bm{\omega}_\bfr/2} \\[3pt]
\frac{1}{m} \left( \bm{q}_{\wfr\bfr} \odot \left(\bm{f}_\text{prop} + \bm{f}_\text{drag}\right)\right) +\bm{g}_\wfr  \\[3pt]
\bm{J}^{-1}\big(\boldsymbol{\tau}_\text{prop}- \boldsymbol\omega_\bfr \times \bm{J}\boldsymbol\omega_\bfr \big) \\[3pt]
\frac{1}{k_\text{mot}} \big(\boldsymbol\Omega_\text{cmd} - \boldsymbol\Omega \big)
\end{bmatrix} \; ,
\end{align}
where $\bm{g}_\wfr= [0, 0, \SI{-9.81}{\meter\per\second^2}]^\intercal$ denotes earth's gravity, $\bm{f}_\text{prop}$, $\boldsymbol{\tau}_\text{prop}$ are the collective force and the torque produced by the propellers, and $\bm{f}_\text{drag}$ is a linear drag term.
The quantities are calculated as follows:
\begin{align}
   \bm{f}_\text{prop} &= \sum_i \bm{f}_i\;, \quad   
   \boldsymbol{\tau}_\text{prop} = \sum_i \boldsymbol{\tau}_i + \bm{r}_{\text{P},i} \times \bm{f}_i \;, \\
   \bm{f}_\text{drag} &= - \begin{bmatrix} k_{vx} v_{\bfr,x}\quad k_{vy} v_{\bfr,y} \quad  k_{vz} v_{\bfr,z} \end{bmatrix}^\top,
\end{align}
where $\bm{r}_{\text{P},i}$ is the location of propeller $i$ expressed in the body frame , $\bm{f}_i$, $\boldsymbol{\tau}_i$ are the forces and torques generated by the $i$-th propeller, and ($k_{vx}$, $k_{vy}$, $k_{vz}$)~\cite{furrer2016rotors, faessler2017differential} are linear drag coefficients.
A commonly used \cite{furrer2016rotors, shah2018airsim} model for the forces and torques exerted by a single propeller is presented in the following: the thrust and drag torque are assumed to be proportional to the square of the propellers' rotational speed. 
The corresponding thrust and drag coefficients $c_l$ and $c_d$ can be readily identified on a static propeller test stand.
By also measuring the rotational speed of the propeller during those tests, the motor time constant $k_\text{mot}$ can be estimated.
\rev{Overall, the force and torque produced by a single propeller are modeled as follows:}
\begin{equation}
\begin{aligned}
    \bm{f}_i(\Omega) = \begin{bmatrix} 0 ~~ 0 ~~ c_{\text{l}}\cdot\Omega^2  \end{bmatrix}^\top\;,&& \bm{\tau}_i(\Omega) = \mat{0 ~~ 0 ~~ c_{\text{d}}\cdot\Omega^2}^\top
    \end{aligned}
\end{equation}

The dynamics are integrated using a symplectic Euler scheme with step size $\SI{1}{\milli\second}$.
For numerical values of the identified mass, inertia, and thrust and drag constants, we refer the reader to Section~\ref{sec:training_details}.

\section{Methodology}\label{sec:methodology}

We address the challenge of robust and agile quadrotor flight using learned control policies by identifying 
\rev{the best choice of action space}
for the task.
We train deep neural control policies that directly map observations $\bm{o}_t$ in the form of platform state and a reference trajectory to control actions $\bm{u}_t$. 
The control policies are trained using model-free reinforcement learning (PPO~\cite{schulman2017proximal}) purely in simulation on a set of over 600~reference trajectories that cover the full performance envelope of the quadrotor. 
We train policies of three different types that only differ in their 
\rev{choice of action space}~$\bm{u}_t$, 
as illustrated in Figure~\ref{fig:overview}: 
\begin{enumerate}
    \item \textbf{Linear Velocity \& Yaw Rate (LV):} 
    \rev{Each action specifies a desired} linear velocity and yaw rate, which are then tracked by a full control stack with access to accurate state estimation.
    $\pi_\text{LV} (\bm{o}_t) \Rightarrow \bm{u}_t = \lbrace v_x, v_y, v_z, \omega_z\rbrace$
    
    \item \textbf{Collective Thrust \& Bodyrate (CTBR):} Each action represents desired collective thrust and bodyrates, which are tracked by a low-level controller using measurements from an inertial sensor.
    $\pi_\text{CTBR} (\bm{o}_t) \Rightarrow \bm{u}_t = \lbrace c, \omega_x, \omega_y, \omega_z \rbrace$
    
    \item \textbf{Single-Rotor Thrust (SRT):} Each action directly specifies desired individual rotor thrusts, which are then applied for the duration of a control step.
    \rev{$\pi_\text{SRT} (\bm{o}_t) \Rightarrow \bm{u}_t = \lbrace f_1, f_2, f_3, f_4 \rbrace$}
\end{enumerate}

All policy types feature a 4-dimensional action space, are fed the same observations $\bm{o}_t$, and are represented by the same network architecture.

\begin{table}[b!]
\vspace*{-6pt}
\caption{\textnormal{Input features to the policy and value networks. \rev{The state is represented by a sliding window of length $H$ of current and previous states, the reference is represented by a receding-horizon window of length $R$ of current and future reference states.} Both networks observe the same state and reference, but only the value network observes privileged information, such as biases in mass, inertia, drag and gravity applied during training with domain randomization.}}
\label{tab:network_inputs}
\vspace*{0pt}
\centering
\begin{tabularx}{1.0\linewidth}{cl|XlX}
\toprule
Input & Components & Dimensions & Policy NW & Value NW \\
\midrule
\multirow{5}{1cm}{State} &z-Position       & $H \times 1$ &  $\checkmark$  &  $\checkmark$ \\
&Velocity       &$H \times 3$ &  $\checkmark$  &  $\checkmark$ \\
&Attitude     &$H \times 9$ &  $\checkmark$  &  $\checkmark$ \\
&Bodyrates       &$H \times 3$ &  $\checkmark$  &  $\checkmark$ \\
&Privileged Info.       &$H \times 7$ &  \xmark  &  $\checkmark$ \\
\midrule
\multirow{4}{1cm}{Reference} & Position       & $R \times 3$ &  $\checkmark$  &  $\checkmark$ \\
&Velocity       &$R \times 3$ &  $\checkmark$  &  $\checkmark$ \\
&Attitude        &$R \times 9$ &  $\checkmark$  &  $\checkmark$ \\
&Bodyrates       &$R \times 3$ &  $\checkmark$  &  $\checkmark$\\
\bottomrule
\end{tabularx}
\vspace*{-12pt}
\end{table}

\subsection{Observations, Actions, and Rewards}
An observation~$\bm{o}_t$ obtained from the environment at time~$t$ consist of (i)~a history of previous states and applied actions and (ii)~the future reference along the trajectory. 
Specifically, the state information contains a history of length~${H=10}$ of the $z$-position of the platform, its velocity, attitude represented as rotation matrix, and bodyrates. 
\rev{Even though the simulator internally uses quaternions, we pass attitude as rotation matrix to the networks to avoid discontinuities~\cite{zhou2019continuity}.}
The reference information consists of a sequence of length~${R=10}$ of future relative position, velocity, and bodyrates as well as the full rotation matrix of the reference.
The \icra{position and velocity components} of the reference states are expressed as the residual from the current state of the quadrotor.
All observations are normalized before passing them to the networks. 

Since the value network is only used during training time, it can access privileged information about the environment that is not accessible to the policy network. 
Specifically, this privileged information contains the mass and inertia biases applied during randomization, as well as the sampled drag coefficients and the additive gravity bias.
An overview of the observation provided to the policy and value network is given in Table~\ref{tab:network_inputs}.
The value network and the policy network share the same architecture but have different parameters. 
The state and reference information are encoded by two separate fully-connected neural networks with 3 hidden layers with 64 neurons each. 
The encodings are then concatenated and fed to a final multilayer perceptron with two layers of 128 neurons each.

\begin{table}[t!]
    \caption{Physical parameters of the simulation. At the start of each rollout, the parameters are sampled from a uniform distribution around the nominal values with the randomization specified above.}
    \label{tab:phys_params}
    \centering
    \begin{tabularx}{1.0\linewidth}{X|ccc}
    \toprule
    Parameter & Nominal Value & Randomization\\
    \midrule
    Mass [\SI{}{\kilogram}] & 0.768 & $\pm 30\%$ \\
    Inertia [\SI{}{\kilogram\meter\squared}] & [2.5e-3, 2.1e-3, 4.3e-3] & $\pm 30\%$\\
    Gravity [\SI{}{\meter\per\second\squared}] & [0.0, 0.0, -9.81] & $\pm 0.4$\\
    $k_{vx}$ [\SI{}{\newton\second\per\meter}] & 0.3 & $\pm 0.3$\\
    $k_{vy}$ [\SI{}{\newton\second\per\meter}] & 0.3 & $\pm 0.3$\\
    $k_{vz}$ [\SI{}{\newton\second\per\meter}] & 0.15 & $\pm 0.15$\\
    $c_l$ [\SI{}{\newton\per\radian\per\second}] & 1.563e-6 & $\pm 0.0$\\
    $c_d$ [\SI{}{\newton\meter\per\radian\per\second}] & 1.909e-8 & $\pm 0.0$\\
    \bottomrule
    \end{tabularx}
\end{table}

We use a dense shaped reward formulation to learn the task of agile trajectory tracking. 
The reward~$r_t$ at timestep~$t$ is given by 
\begin{align}\label{eq:reward}
    r_t = &- (\bm{x}_t - \bm{x}_{\text{ref},t})^\top \bm{Q} (\bm{x}_t - \bm{x}_{\text{ref},t}) \\\nonumber
    &- (\bm{u}_t - \bm{u}_{\text{ref},t})^\top \bm{R} (\bm{u}_t - \bm{u}_{\text{ref},t}) - r_\text{crash}\; ,
\end{align}
where $\bm{Q}$ and $\bm{R}$ are diagonal matrices, $\bm{x}_t$ the full state of the quadrotor, $\bm{u}_t$ the applied action, $\bm{x}_{\text{ref},t}$ and $\bm{u}_{\text{ref},t}$ their respective references, and $r_\text{crash}$ is a binary penalty that is only active when the altitude of the platform is negative, which also ends the episode.

\subsection{Policy Learning}\label{sec:learning}
\icra{All control policies are} trained using Proximal Policy Optimization~(PPO)~\cite{schulman2017proximal}. 
PPO has been shown to achieve state-of-the-art performance on a set of continuous control tasks and is well suited for learning problems where interaction with the environment is fast.
Data collection is performed by simulating 50 agents in parallel. 
At each environment reset, every agent samples a new trajectory from the set of training trajectories and is initialized with bounded perturbation at the start of the trajectory. 

Inspired by prior work on simulation to reality transfer, we perform randomization of the dynamics of the platform during training time and apply Gaussian noise to the policy observations. 
Specifically, we randomize mass, inertia, aerodynamic drag, and thrust variations of the quadrotor.

\subsection{Training Details}\label{sec:training_details}
The policies are trained in a simulated quadrotor environment implemented using TensorFlow Agents. 
The nominal quadrotor parameters such as mass and inertia are identified from the real platform and are summarized in Table~\ref{tab:phys_params} together with the amount of randomization applied at training time. 
\icra{Training} hyperparameters specified in Table~\ref{tab:ppo_params}.
\begin{table}[t!]
    \caption{Training hyperparameters.}
    \label{tab:ppo_params}
    \centering
    \begin{tabularx}{1\linewidth}{X|c}
    \toprule
    Hyperparameter & Value\\
    \midrule
    $\gamma$ (discount factor)  & 0.98\\
    Actor learning rate & 3e-4 \\
    Critic learning rate & 3e-4 \\
    Entropy regularization & 1e-2 \\
    $\varepsilon$ (importance ratio clipping) & 0.2 \\
    \bottomrule
    \end{tabularx}
    \vspace*{-12pt}
\end{table}

During trajectory tracking, the agent receives at each timestep a reward that penalizes tracking error and deviation from the reference action as laid out in Eq.~\eqref{eq:reward}.
The matrices $\bm{Q}$ and $\bm{R}$ have nonzero elements only on the diagonal. 
Specifically, we use ${\bm{Q} = \text{diag} \lbrace 0.1 \cdot \mathbf{1}_{3\times1},0.02 \cdot \mathbf{1}_{9\times1},0.002 \cdot \mathbf{1}_{3\times1},0.01 \cdot \mathbf{1}_{3\times1}  \rbrace}$ and ${\bm{R} = \text{diag} \lbrace 0.001 \cdot \mathbf{1}_{4\times1}\rbrace}$.
The episode is terminated when the quadrotor crashes (i.e. $p_z \leq 0.0$) with a reward of $r_\text{crash} = -500$.

\section{Experiments}
\label{sec: exp}
We design our experimental setup to investigate the influence of the choice of 
\rev{action space}
on flight performance. 
Specifically, we design our experiments to answer the following research questions:
(i)~How is the peak control performance in situation of perfect model identification affected by the \rev{actuation model?} %
(ii)~How does the choice of 
\rev{action space}
affect the robustness against model mismatch?
\rev{(iii)~What is the impact of the choice of action space on training data requirement?}

We evaluate the performance of all policies on a set of test trajectories of varying agility, spanning from a hover trajectory up to a racing trajectory~\cite{foehn2021time} that requires to perform accelerations beyond 3g to track. 
All test trajectories are within the distribution of training trajectories and are feasible, i.e. they do not exceed the platform limits. Table~\ref{tab:test_trajectories} shows the key metrics of all test trajectories.

\begin{table}[b!]
\vspace*{-12pt}
\caption{\textnormal{Maxima of velocity, mass-normalized collective thrust and bodyrates of the  test trajectories.}}
\label{tab:test_trajectories}
\begin{tabularx}{1\linewidth}{X|ccc}
\toprule
Trajectory & $ \Vert \mathbf v \Vert_\text{max}[\SI{}{\meter\per\second}]$& $ c_\text{max}[\SI{}{\meter\per\second\squared}]$ & $ \Vert \boldsymbol\omega \Vert_\text{max}[\SI{}{\radian\per\second}]$ \\
\midrule
Hover       & 0.0    & 9.81  &  0.0   \\
RandA      & 3.87   & 12.68 &  1.27  \\
RandB      & 6.36   & 13.54 &  1.52  \\
RandC      & 8.92   & 14.52 &  1.93  \\
RaceA    & 10.48  & 16.18 &  5.74  \\
RaceB    & 11.97  & 24.94 &  8.37  \\
Split-S     & 12.40  & 26.35 &  6.11  \\
RaceC    & 14.22  & 33.04 &  11.56  \\
\bottomrule
\end{tabularx}

\end{table}

\subsection{Simulation Experiments}
In a set of controlled experiments in simulation, the tracking performance of each policy is investigated. 
We compare performance with respect to \icra{average} positional tracking error.
Experiments are performed on the test trajectories in two settings: (i)~in the \textit{Nominal} setting, the test environment perfectly matches the training environment; 
(ii)~in the \textit{Model Mismatch} setting, the environment at test time is different from the training environment. 
Specifically, we use in setting~(ii) a quadrotor simulation that was identified from real flight data and uses blade-element momentum theory to accurately model the aerodynamic forces acting on the platform~\cite{bauersfeld2021neurobem}. 
We also apply a control delay of \SI{20}{\milli\second} to simulate wireless communication latency. 
Note that we can only use this simulation at test time, since it is computationally too expensive to run it at training time.\\
While setting~(i) is focused on the maximum possible performance achievable by a method and its training data requirement, setting~(ii) investigates the robustness of policies against model mismatch.
All policies tested in setting~(i) have been trained specifically for the nominal environment without any randomization, while the policies tested in setting~(ii) have been trained on a distribution of environments as explained in Section~\ref{sec:learning}.
We also compare against two state-of-the-art classical control approaches: MPC-SRT represents an optimization-based controller~\cite{torrente2021data} that directly controls at individual rotor thrust level, while MPC-CTBR makes use of a low-level controller.
All learned policies are run at a constant frequency of \SI{50}{\hertz}, while the traditional controllers are executed at \SI{100}{\hertz}.

\begin{table}[t!]
\caption{\textnormal{\icra{Average} positional tracking error in centimeter on each test trajectory in case of no model mismatch. The table reports results for learned policies (SRT, CTBR, LV), and traditional approaches (MPC-SRT, MPC-CTBR). \rev{Results report mean and standard deviation for 10 trained policies.}}}
\label{tab:pos_tracking_vanilla}
\footnotesize
\setlength{\tabcolsep}{2pt}
\begin{tabularx}{1\linewidth}{X|ccc|cc}
\toprule
 & SRT & CTBR & LV & MPC-SRT & MPC-CTBR \\
 \midrule
Hover &   1.0$\pm$0.2 & \textbf{0.6$\pm$0.2} & 7.0$\pm$1.6 & \textbf{0.1} & 0.2 \\  
RandA & 1.5$\pm$0.2 & \textbf{0.9$\pm$0.1} & 15.4$\pm$3.0 &  \textbf{0.2} & 0.3 \\  
RandB & 2.4$\pm$0.2 & \textbf{1.6$\pm$0.1} & 61.5$\pm$21.0 &  \textbf{0.2} & 0.4 \\ 
RandC & 3.0$\pm$0.3 & \textbf{2.0$\pm$0.2} & 85.7$\pm$11.5 & \textbf{0.2} & 0.4 \\ 
RaceA & \textbf{5.0$\pm$1.2} & \textbf{5.0$\pm$1.0} & 121.1$\pm$25.8 & \textbf{0.3} & 1.3 \\ 
RaceB & 7.1$\pm$1.8 & \textbf{6.9$\pm$1.5} & 170.2$\pm$16.3 & \textbf{0.7} & 3.0 \\ 
Split-S & \textbf{3.5$\pm$0.4} & 6.6$\pm$1.1 & 92.1$\pm$20.8 &  \textbf{1.0} & 2.1 \\ 
RaceC & \textbf{9.2$\pm$3.2} & 12.3$\pm$2.2 & 197.9$\pm$38.1 &\textbf{1.2} & 3.9  \\
\bottomrule
\end{tabularx}
\vspace*{-12pt}
\end{table}

\rev{
\mypara{Nominal Model} Table~\ref{tab:pos_tracking_vanilla} shows the results of the experiments in the \textit{Nominal} setting (i). 
SRT and CTBR policies perform comparable in this setting, with CTBR marginally outperforming on slower trajectories, while SRT performs slightly better on the more aggressive maneuvers.
These results confirm previous findings from experiments in the domain of 2D locomotion~\cite{peng2017learning}: policies that operate in concert with an underlying low-level controller outperform end-to-end policies. 
The policies that produce linear velocity commands (LV) perform inferior especially for agile maneuvers.
This can be explained by the fact that the action space of linear velocity commands does not correctly represent the dynamic constraints of a quadrotor platform, which leads to a reduced maneuverability. 
This result extends the findings of~\cite{peng2017learning} and shows that more abstraction does not necessarily lead to better performance.
Compared to the learned policies, the traditional control approaches (MPC-SRT, MPC-CTBR) perform significantly better in the \textit{Nominal} setting. This results is expected, as the system dynamics implemented in the MPC exactly match the simulated dynamics of the platform.
We still provide these results to allow a comparison with traditional control approaches.

}

\rev{
\mypara{Model Mismatch} Table~\ref{tab:pos_tracking_bem_20_latency} shows the results of the \textit{Model Mismatch} scenario. Controllers that directly specify single rotor thrusts exhibit a significant reduction in performance, especially for agile trajectories: SRT has a significantly higher tracking error for slow trajectories and often crashes on the faster maneuvers; MPC-SRT also has higher tracking error and even crashes on RaceB and RaceC.
We report \textit{crash}, as soon as one policy crashes on the maneuver.
The CTBR policies (as well as MPC-CTBR) are less affected by the model mismatch and can still execute all maneuvers with a modest increase in tracking errors. 
The LV policies show a smaller sensitivity to the model mismatch, but are still consistently outperformed by the CTBR policies on all trajectories.
}

\begin{table}[t!]
\centering
\caption{\textnormal{\icra{Average} positional tracking error in centimeter on each test trajectory obtained in a quadrotor simulator based on blade-element momentum theory with a control delay of $\SI{20}{\milli\second}$. \rev{Results report mean and standard deviation for 10 trained policies.}}}
\label{tab:pos_tracking_bem_20_latency}
\setlength{\tabcolsep}{2pt}
\begin{tabularx}{1\linewidth}{X|ccc|cc}
\toprule
 & SRT & CTBR & LV & MPC-SRT & MPC-CTBR \\
 \midrule
Hover &  11.3$\pm$4.5 &  \textbf{0.6$\pm$0.5} & 6.7$\pm$2.0 & 1.0 &  \textbf{0.5} \\
RandA & 12.0$\pm$4.0 &  \textbf{1.2$\pm$0.5} & 17.8$\pm$1.4 & 3.3 &  \textbf{1.1} \\ 
RandB &  14.4$\pm$2.4 &  \textbf{2.2$\pm$0.8} & 57.0$\pm$12.0 & 7.0 &  \textbf{2.0} \\  
RandC &  17.6$\pm$5.9 & \textbf{2.6$\pm$0.8} & 78.9$\pm$13.4 & 8.5 & \textbf{2.6} \\ 
RaceA &  crash & \textbf{5.6$\pm$1.7} & 144.0$\pm$20.1 & 12.6 &  \textbf{4.8} \\  
RaceB &  crash & \textbf{10.0$\pm$4.0} & 171.4$\pm$17.0 & crash &  \textbf{6.3} \\  
Split-S &  crash & \textbf{6.9$\pm$2.6} & 83.8$\pm$9.7 & \textbf{11.3} &  11.4 \\  
RaceC & crash & \textbf{14.9$\pm$5.5} & 176.7$\pm$22.0 & crash & \textbf{7.5} \\
\bottomrule
\end{tabularx}
\vspace*{-12pt}
\end{table}

\begin{figure*}[t!]
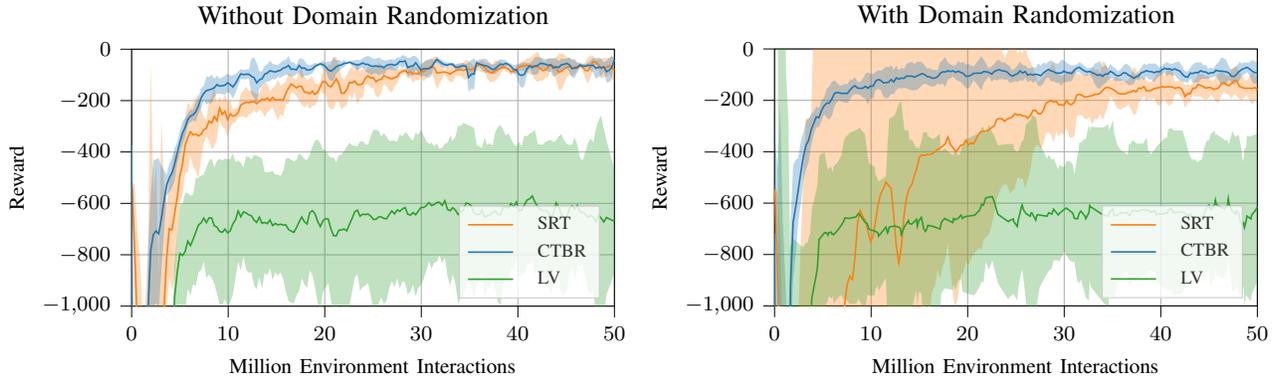

    \centering
    \input{figures/learning_curve_vanilla}%
    \input{figures/learning_curve_randomized}%
    \caption{\rev{Learning curves of policies trained without (left) and with (right) domain randomization. All policies are trained for a total of 50M environment interactions. Learning curves show mean performance and standard deviation computed over all trained policies.}}
    \label{fig:learning_curves}
\end{figure*}

\rev{\mypara{Training Data Requirement} Figure~\ref{fig:learning_curves} depicts the learning curves of all policies in case of no domain randomization (left) and with domain randomization (right). 
All policies have been trained for a total of 50M environment interactions.
The learning curves also show the robustness of CTBR and LV to changes in the platform dynamics, we even observed that training with a randomized platform accelerates learning in the early stages of training. 
In contrast, the learning curves of SRT in case of domain randomization initially exhibit a high variance, train slower, and converge to a final performance \icra{substantially} lower than in case of no domain randomization.
}

\begin{figure*}
\vspace{9pt}
\begin{tikzpicture}
\path (0,0) -- (\linewidth,0);
\draw (1,0) rectangle (\linewidth-0.5cm,0.5);
\end{tikzpicture}
\vspace*{-1.4cm}

\begin{tikzpicture}[font=\scriptsize]
\begin{axis}[
width = 0.33\linewidth,
height = 0.21\linewidth,
xlabel = {Latency [ms]},
ylabel = {Position RMSE [m]},
title = {\bf Influence of Latency - Hover},
title style={yshift=3ex},
xmin = 0,
xmax = 100,
ymin = 0,
ymax = 0.25,
xmajorgrids,
ymajorgrids,
xlabel shift = -3pt,
ylabel shift = -3pt,
legend style={align=left, at={(axis cs:90, 0.261)}, anchor=south, draw=none, fill=none},
]

\addplot [color=color1,thick, dashed, mark=*, mark options={solid} ] table [row sep=crcr] {
0       0.004 \\
20    0.004 \\
40    0.004 \\
60    0.006 \\
80    0.007 \\
100 0.054 \\
};
\addlegendentry{CTBR}

\addplot [color=color2, thick, dashed, mark=*, mark options={solid}] table [row sep=crcr] {
0       0.021 \\
20    0.106 \\
40    0.25 \\
60    0.25 \\
80    0.25 \\
100 0.25 \\
};
\addplot [color=color3, thick, dashed, mark=*, mark options={solid}] table [row sep=crcr] {
0   0.005 \\
20  0.005 \\
40  0.005 \\
60  0.005 \\
80  0.03 \\
100 0.25 \\
};
\addplot [color=color4, thick, dashed, mark=*, mark options={solid}] table [row sep=crcr] {
0   0.087 \\
20  0.087 \\
40  0.087 \\
60  0.086 \\
80  0.097 \\
100 0.112 \\
};
\end{axis}
\end{tikzpicture}
\hfill
\begin{tikzpicture}[font=\scriptsize]
\begin{axis}[
width = 0.33\linewidth,
height = 0.21\linewidth,
xlabel = {Latency [ms]},
ylabel = {Position RMSE [m]},
title = {\bf Influence of Latency - RandomC},
title style={yshift=3ex},
xmin = 0,
xmax = 100,
ymin = 0,
ymax = 1.5,
xmajorgrids,
ymajorgrids,
xlabel shift = -3pt,
ylabel shift = -3pt,
legend style={align=left, at={(axis cs:0, 1.575)}, anchor=south west, draw=none, fill=none, text width=5.0em},
legend columns=3,
]
\addplot [color=color1, thick, dashed, mark=*, mark options={solid}, forget plot ] table [row sep=crcr] {
0       0.034 \\
20    0.034 \\
40    0.034 \\
60    0.044 \\
80    0.081 \\
100 1.0  \\ %
};

\addplot [color=color2, thick, dashed, mark=*, mark options={solid}] table [row sep=crcr] {
0       0.036 \\
20    0.164 \\
40    1.5 \\
60    1.5 \\
80    1.5 \\
100 1.5 \\
};
\addlegendentry{SRT}

\addplot [color=color3, thick, dashed, mark=*, mark options={solid}, forget plot] table [row sep=crcr] {
0   0.025 \\
20  0.026 \\
40  0.026 \\
60  0.027 \\
80  0.032 \\
100 0.903 \\
};

\addplot [color=color4, thick, dashed, mark=*, mark options={solid}] table [row sep=crcr] {
0   0.638 \\
20  0.638 \\
40  0.602 \\
60  0.696 \\
80  0.767 \\
100 0.825 \\
};
\addlegendentry{LV}
\end{axis}
\end{tikzpicture}
\hspace*{-0.2cm}
\begin{tikzpicture}[font=\scriptsize]
\begin{axis}[
width = 0.33\linewidth,
height = 0.21\linewidth,
xlabel = {Latency [ms]},
ylabel = {Position RMSE [m]},
title = {\bf Influence of Latency - RaceC},
title style={yshift=3ex},
xmin = 0,
xmax = 100,
ymin = 0,
ymax = 1.5,
xmajorgrids,
ymajorgrids,
xlabel shift = -3pt,
ylabel shift = -3pt,
legend style={align=left, at={(axis cs:0, 1.575)}, anchor=south west, draw=none, fill=none},
]
\addplot [color=color1, thick, dashed, mark=*, mark options={solid}, forget plot ] table [row sep=crcr] {
0       0.110 \\
20    0.111 \\
40    0.324 \\
60    0.644 \\
80    1.0 \\ %
100 1.0 \\ %
};

\addplot [color=color2, thick, dashed, mark=*, mark options={solid}, forget plot] table [row sep=crcr] {
0       1.5 \\  %
20    1.5 \\
40    1.5 \\
60    1.5 \\
80    1.5 \\
100 1.5 \\
};

\addplot [color=color3, thick, dashed, mark=*, mark options={solid}] table [row sep=crcr] {
0   0.075 \\
20  0.074 \\
40  0.076 \\
60  1.143 \\
80  1.5 \\ %
100 1.5 \\ %
};
\addlegendentry{MPC-CTBR}

\addplot [color=color4, thick, dashed, mark=*, mark options={solid}] table [row sep=crcr] {
0   1.372 \\
20  1.372 \\
40  1.380 \\
60  1.394 \\
80  1.420 \\
100 1.548 \\
};
\end{axis}
\end{tikzpicture}
\caption{Sensitivity to control delay on three trajectories of increasing agility. The results show that policies that operate on single rotor thrust (SRT) are less robust against control delay. }
\vspace{-12pt}
\label{fig:latency_plots}
\end{figure*}
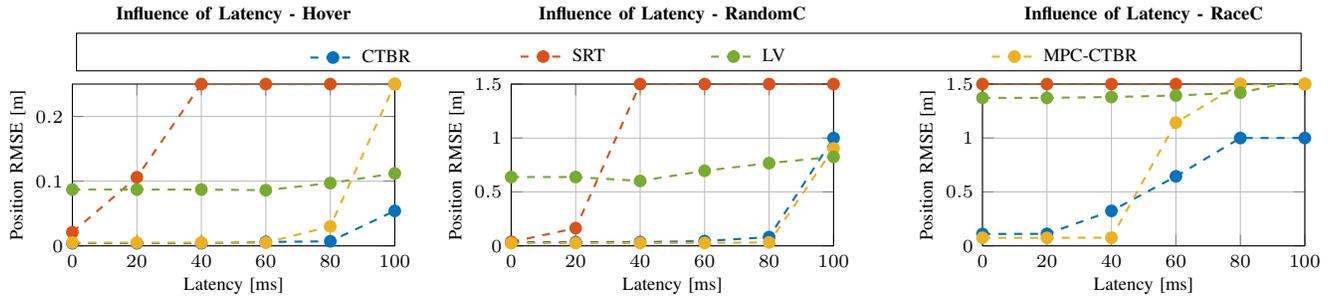

\mypara{Influence of Delay}
Our experiments show that the performance of the tested control policies varied significantly in case of unknown control delay. 
Figure~\ref{fig:latency_plots} shows that policies that operate at higher abstraction levels such as LV or CTBR are less sensitive to such delay. 
Furthermore, accurate identification of control delay is more important for agile trajectories; while hover is possible for CTBR without a noticeable decrease in performance for latencies up to \SI{60}{\milli\second}, the same latency leads to a crash on the racing trajectory.

\begin{table}[b!]
\vspace{-12pt}
\caption{\textnormal{Positional tracking error in centimeter on a set of test trajectories executed in the real world. \rev{Results report mean and standard deviation for 5 trained policies.}}}
\label{tab:pos_tracking_real_world}
\footnotesize
\begin{tabularx}{1\linewidth}{X|cc|c}
\toprule
 & CTBR & LV & MPC-CTBR \\
\midrule
Hover & \textbf{4.4$\pm$1.4} & 6.2$\pm$2.0 & 3.0 \\  
RandA & \textbf{8.1$\pm$1.0} & 60.0$\pm$16.8 & 8.0 \\  
RandB & \textbf{8.6$\pm$0.8} & 87.0$\pm$30.3 & 8.0 \\  
RandC & \textbf{47.8$\pm$9.9} & 134.8$\pm$19.6 & 14.0 \\ 
Circle & \textbf{31.8$\pm$4.4} & 170.7$\pm$11.6 & 25.0 \\  
Lemniscate & \textbf{26.8$\pm$4.4} & 189.5$\pm$13.7 & 16.0 \\  
Racing & \textbf{53.0$\pm$9.2} & 200.8$\pm$14.5 & 20.0 \\
\bottomrule
\end{tabularx}
\end{table}

\subsection{Real World Experiments}
We assess the performance of different control policies when deployed on a real quadrotor platform. 
As in the simulation experiments, we execute a set of trajectories and compare tracking performance between the methods presented in Section~\ref{sec:methodology}.
We encourage the reader to watch the supplementary video to understand the dynamic nature of these experiments.

The results of the real world experiments are shown in Table~\ref{tab:pos_tracking_real_world}. 
Due to its significant sensitivity to control delays and a communication latency of \SI{60}{\milli\second} imposed by the real system, the SRT policies could not be deployed. 
The CTBR policies instead manage to fly unseen maneuvers on the real platform despite the control delay. 
The LV policies transfer to the real platform as well, but CTBR significantly outperforms on agile trajectories. 
\icra{Compared to the results in the BEM simulator, tracking errors are higher in the real world mainly due to unmodelled effects such as varying battery voltage, imperfect motor thrust mappings, and torque imbalances due to imperfect mass distribution.}
Throughout the tested trajectories, the CTBR policies reach accelerations of up to 3g and speeds beyond \SI{45}{\kilo\meter\per\hour}, which outperforms the previous state of the art in learning-based quadrotor control by a factor of 3 in terms of speed.

\section{Conclusion}
\label{sec: conclusion}
We presented a comparison of learning-based controllers for agile quadrotor flight.
We compared policies that \rev{specify} individual rotor thrusts, collective thrust and bodyrates, and linear velocity commands. 
While all tested policy types were able to learn a universal flight controller, they differed strongly in terms of peak performance and robustness against dynamics mismatch. 
We identified that policies \rev{producing} collective thrust and bodyrates exhibit strong resilience against dynamics mismatch and transfer well between domains while retaining high agility. 
This work can serve as guideline for future work on learning-based quadrotor control by identifying the control input modality that is best suited for agile flight.

{\footnotesize
\balance

}

\end{document}